\documentclass[10pt,twocolumn,letterpaper]{article}

\usepackage{cvpr}
\usepackage{times}
\usepackage{epsfig}
\usepackage{graphicx}
\usepackage{amsmath}
\usepackage{amssymb}
\usepackage{multirow}
\usepackage{color}
\usepackage{array}
\usepackage{bigints}
\usepackage[dvipsnames]{xcolor}
\usepackage{booktabs}

\usepackage[pagebackref=true,breaklinks=true,letterpaper=true,colorlinks,bookmarks=false]{hyperref}

\cvprfinalcopy 

\begin{document}

\title{Action Recognition from Single Timestamp Supervision in Untrimmed Videos}

\author{Davide Moltisanti\\Visual Information Lab\\
University of Bristol\\
{\tt\small davide.moltisanti@bristol.ac.uk}
\and
Sanja Fidler\\
University of Toronto\\NVIDIA \hspace{12pt}Vector Institute\\\
{\tt\small fidler@cs.toronto.edu}
\and
Dima Damen\\Visual Information Lab\\
University of Bristol\\
{\tt\small dima.damen@bristol.ac.uk}
}
\maketitle

\begin{abstract}
Recognising actions in videos relies on labelled supervision during training, typically the start and end times of each action instance. This supervision is not only subjective, but also expensive to acquire. 
Weak video-level supervision has been successfully exploited for recognition in untrimmed videos, however it is challenged when the number of different actions in training videos increases.  
We propose a method that is supervised by single timestamps located around each action instance, in untrimmed videos. 
We replace expensive action bounds with sampling distributions initialised from these timestamps.
We then use the classifier's response to iteratively update the sampling distributions.
We demonstrate that these distributions converge to the location and extent of discriminative action segments.

We evaluate our method on three datasets for fine-grained
recognition, with increasing number of different actions
per video, and show that single timestamps offer a
reasonable compromise between recognition performance
and labelling effort, performing comparably to full temporal
supervision. Our update method improves top-1 test
accuracy by up to 5.4\%. across the evaluated datasets.

\end{abstract}

\section{Introduction}
\label{sec:introduction}

Typical approaches for action recognition in videos rely on full temporal supervision, i.e. on the availability of the action start and end times for training. When the action boundaries are available, all (or most of) the frames enclosed by the temporal bounds can be considered relevant to the action, and thus state-of-the-art methods randomly or uniformly select frames to represent the action and train a classifier~\cite{simonyan2014two,feichtenhofer2016convolutional, wang2016temporal, carreira2017quo, yeung2016end, kalogeiton2017action}. 
Collecting these boundaries is not only notoriously burdensome and expensive, but also potentially ambiguous and often arbitrary~\cite{moltisanti2017trespassing,Sigurdsson_2017_ICCV,cheron2018flexible}. 

With an increasing need for bigger video datasets, it is important to scale up the annotation process to foster more rapid advance in video understanding.
In this work, we attempt to alleviate such annotation burden, using \textit{single} roughly aligned timestamp annotations in untrimmed videos - i.e. videos labelled with only one timestamp per action, located close to the action of interest. Such labelling is quicker to collect, and importantly is easier to communicate to annotators who do not have to decide when the action starts or ends, but only label one timestamp within or close to the action. Single timestamps can alternatively be collected from audio narrations and video subtitles~\cite{Damen2018EPICKITCHENS,alayrac2016unsupervised}.

\begin{figure*}[t!]
	\centering
	\includegraphics[width=\textwidth]{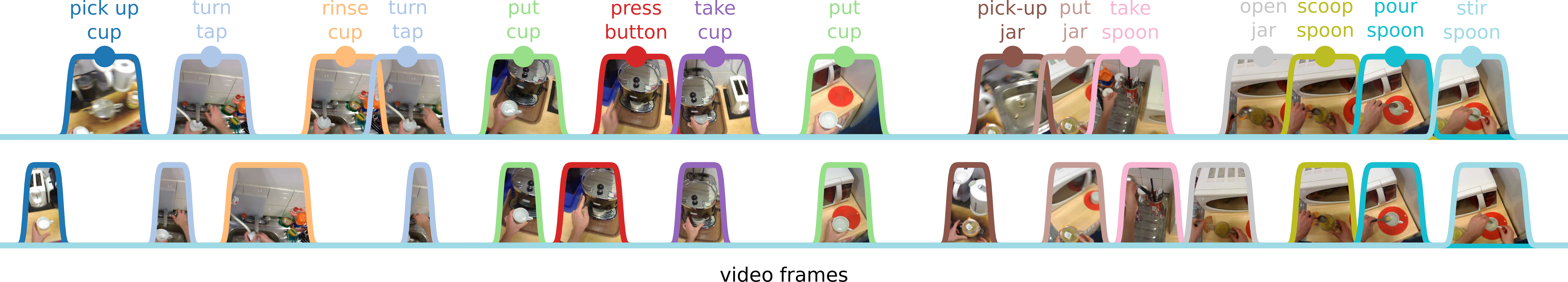}
	\caption{Replacing action boundaries with sampling distributions in an untrimmed video, given single timestamps (coloured dots at the centre of each distribution). The initial distributions (top) may overlap  (e.g. `put jar', `take spoon') and contain background frames. We iteratively refine the distributions (bottom) using the classifier response during training.}
	\label{fig:initialGv}
\end{figure*}

To utilise this weak supervision, we propose a sampling distribution, initialised from the single timestamps, to select relevant frames to train an action recognition classifier. 
Due to the potential coarse location of the timestamps, and to actions having different lengths, the initial sampling distributions may not be well aligned with the actions, as showed in Figure~\ref{fig:initialGv} (top).
We thus propose a method to update the parameters of the sampling distributions during training, using the classifier's response, in order to sample more relevant frames and reinforce the classifier (Figure~\ref{fig:initialGv}, bottom). 

Our attempt is inspired by similar approaches for single point annotations in image based semantic segmentation~\cite{bearman2016s}, where results achieved using such point supervision have slightly lower accuracy than those obtained with fully annotated masks, but outperform results obtained with image-level annotations. Correspondingly, we show that single timestamp supervision for action recognition outperforms video-level supervision. 

We test our method on three datasets~\cite{jiang2014thumos,Damen2014a,Damen2018EPICKITCHENS}, of which~\cite{Damen2018EPICKITCHENS} is annotated with single timestamps from live audio commentary. 
We show that our update method converges to the location and temporal extent of actions in the three datasets, and boosts initial accuracy on the three datasets.
We additionally demonstrate the advantages of curriculum learning during this update process, and the robustness of our approach to the initial parameters of the sampling distributions.
When single timestamps are consistently within the action boundaries, our approach is comparable to strongly supervised models on all datasets.

\section{Related Work}

We review recent works using weak temporal labels for action recognition and localisation. For a review of works that use strong supervision, we refer the reader to~\cite{herath2017going}. We divide the section into works using video-level, transcript and point-level supervision.

\vspace{-10pt}

\paragraph{Video-level supervision} provides the weakest cue, signalling only the presence or absence of an action in an untrimmed video, discarding any temporal ordering.
When only a few different actions are present in an untrimmed video, video-level supervision can prove sufficient to learn the actions even from long videos, as recently shown in~\cite{wang2017untrimmednets,nguyen2017weakly,singh2017hide,shou2018autoloc,paul2018w}. In these works, the authors use such supervision to train a model for action classification and localisation, achieving results often comparable to those obtained with strongly supervised approaches. 
However, all these works evaluate their approach on the THUMOS 14~\cite{jiang2014thumos} and Activity Net~\cite{caba2015activitynet} datasets, which contain mainly one class per training video. 
In this work, we show that as the number of different actions per training video increases, video-level labels do not offer sufficient supervision.

\vspace{-10pt}

\paragraph{Transcript supervision} offers an ordered list of action labels in untrimmed videos, without any temporal annotations~\cite{bojanowski2014weakly,bojanowski2015weakly,huang2016connectionist,richard2016temporal,kuehne2017weakly,richard2017weakly,richard2018neuralnetwork,ding2018weakly}. 
Some works~\cite{ding2018weakly,kuehne2017weakly,richard2016temporal} assume the transcript includes knowledge of `background', specifying whether the actions occur in succession or with gaps.
In~\cite{ding2018weakly}, uniform sampling of the video is followed by iterative refinement of the action boundaries. The refinement uses the pairwise comparison of softmax scores for class labels around each boundary, along with linear interpolation. This iterative boundary refinement strategy is conceptually similar to ours. However, the approach in~\cite{ding2018weakly} assumes no gaps are allowed between neighbouring actions. This requires knowledge of background labels in order for the method to operate. 

\vspace{-10pt}

\paragraph{Point-level supervision} refers to using a single pixel or a single frame as a form of supervision.
This was attempted for semantic segmentation, by annotating single points in static images~\cite{bearman2016s} and subsequently used for videos~\cite{mettes2016spot,cheron2018flexible}.
In~\cite{mettes2016spot} a single pixel is used to annotate the action, in a subset of frames, both spatially and temporally. When combining this weak supervision with action proposals, the authors show that it is possible to achieve comparable results to those obtained with full and much more expensive per-frame bounding boxes. 
More recently, several forms of weak supervision, including single temporal points, are evaluated in~\cite{cheron2018flexible} for the task of spatio-temporal action localisation. This work uses an off-the-shelf human detector to extract human tracks from the videos, integrating these with the various annotations in a unified framework based on discriminative clustering. 

In this work, we also use a single temporal point per action for fine-grained recognition in videos. However, unlike the works above~\cite{mettes2016spot,cheron2018flexible} which consider the given annotations correct, we actively refine the temporal scope of the given supervision, under the assumption that the given annotated points may be misaligned with the actions and thus lead to incorrect supervision. 
We show this to effectively converge, when tested on three datasets with varying complexity, in the number of different actions in untrimmed training videos. We detail our method next.

\begin{figure*}[t!]
	\centering
	\includegraphics[width=\textwidth]{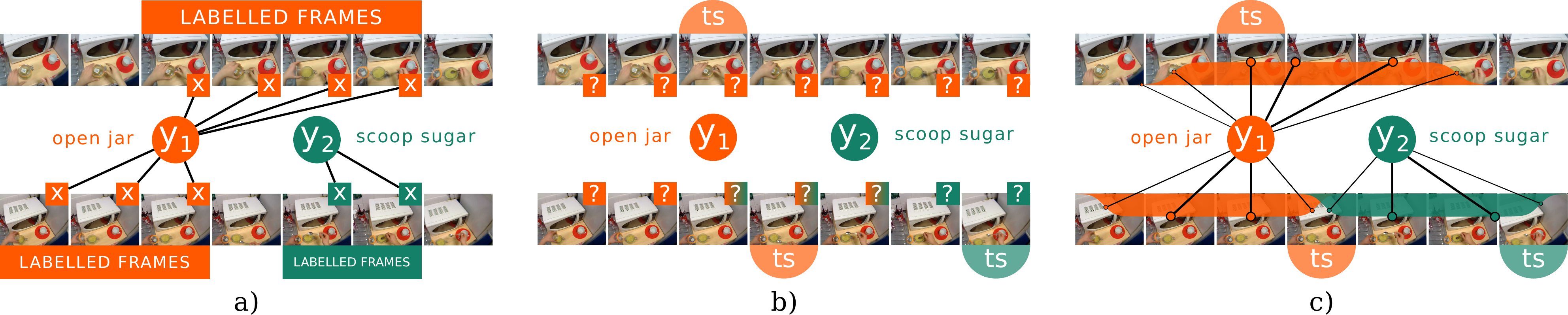}
	\caption{When start/end times are available (a), all frames within labelled boundaries can be assigned to the class label. Since action bounds are not available (b), our method aims to iteratively update the mapping between frames and class labels (c). Top and bottom plots depict different videos.
	}
	\label{fig:mapping}
\end{figure*}

\section{Recognition from Single Timestamp Supervision}
\label{sec:method}

In this work, we consider the case where a set of untrimmed videos, containing multiple different actions, are provided for the task of fine-grained action recognition. 
That is the task of training a classifier $f(x) = y$ that takes a frame (or a set of frames) $x$ as input to recognise a class $y$ from the visual content of $x$. 
Our method is classifier agnostic, i.e. we do not make any assumptions about the nature of the classifier.

The typical annotation for this task is given by the actions' start and end times, which delimit the temporal scope of each action in the untrimmed video, as well as the class labels. We refer to this labelling as \textit{temporal bounds annotation}. When using this supervision, the classifier can be trained using frames between the corresponding start/end timestamps. 
When replacing these annotations with a single timestamp per action instance,
training a classifier is not straightforward. 
Figure~\ref{fig:mapping} compares temporal bounds (a) to the single timestamp annotations (b).
In Figure~\ref{fig:mapping}b, it is not evident which frames could be used to train the classifier when only roughly aligned single timestamps are available. 
While being close to the action, the frame corresponding to the single timestamp could represent the background or another action. Additionally, the extent of the action is unknown. Our method is based on the reasonable assumption that multiple instances of each class have been labelled, allowing the model to converge to the correct frames.

We propose a sampling distribution (Section~\ref{sec:sample}) to select training frames for a classifier starting from the annotated timestamps, as depicted in Figure~\ref{fig:mapping}c. After initialisation (Section~\ref{sec:initialisation}), we iteratively update the parameters of the sampling distributions based on the classifier's response, in the attempt to correct misplaced timestamps and reinforce the classifier with more relevant frames (Section~\ref{sec:update}).

\subsection{Sampling Distribution}
\label{sec:sample}

We propose to replace the unavailable action bounds with a sampling distribution that can be used to select frames for training a classifier. 
For simplicity, we assume here our classifier is frame-based and takes as input a single frame. We relax this assumption later.

We argue that the sampling distribution should resemble the output of a strong classifier, i.e. a plateau of high classification scores for consecutive frames containing the action, with low response elsewhere. 
Another desirable property of this function is differentiability, so that it can be learnt or tuned. The Gaussian probability density function (pdf) is commonly used to model likelihoods, however it does not exhibit a plateau response, peaking instead around the mean and steadily dropping from the peak. The gate function by definition exhibits a sharp plateau, however it is not differentiable. 
We propose the following function to model the probability density of the sampling distributions:
\begin{equation}
	g(x\ |\ c,w,s) = \frac{1}{(e^{s(x-c-w)} + 1) (e^{s(-x+c-w)} + 1)}
	\label{eq:plateauFunction}
\end{equation}
The parameter $c$ models the centre of the plateau, while $w$ and $s$ model respectively its width (equal to $2w$) and the steepness of its side slopes. The range of the function is~$[0, 1]$.
In our setting, $g$ is defined over the frames $x$ of an untrimmed video.
We refer to $g$ as the \textbf{plateau function} for the remainder of the text. 

\subsection{Initialising the Model}
\label{sec:initialisation}

We initialise the sampling distributions from the single timestamp annotations.
Let $a_i^v$ be the i-\textit{th} single timestamp in an untrimmed video $v$ and let $y_i^v$ be its corresponding class label, with ${i \in \{1..N_v\}}$ and ${v \in \{1..M\}}$. 
For each $a_i^v$, we initialise a sampling distribution centred on the timestamp, with default parameters $w$ and $s$.
We denote the parameters of the corresponding sampling distribution with ${\beta_i^v = (c_i^v, w_i^v, s_i^v)}$, where $c_i^v = a_i^v$, and accordingly we denote the corresponding sampling distribution with $G(\beta_i^v)$. We will use $G(\beta_i^v)$ to sample training frames for the class indicated by $y_i^v$.  

Note that, due to the close proximity of some timestamps, the initialised plateaus may overlap considerably~(Figure~\ref{fig:initialGv}, top). 
We could decrease the overlap by shrinking the plateaus. However, given that we do not know the temporal extent of the actions, this may result in missing important frames. 
We choose to allow the overlap, and set $w$ and $s$ to default values that give all actions the chance to be learnt from the same number of frames. 

Frames sampled from these distributions might be background frames, or be associated with incorrect action labels. 
To decrease noise, we rank frames sampled from all untrimmed videos based on the classifier's response, and select the most confident frames for training, inspired by curriculum learning~\cite{bengio2009curriculum}. Let $P(k|x)$ denote the softmax scores of a frame $x$ for a class $k$. Let: 
\begin{equation}
\resizebox{\columnwidth}{!}{$\begin{aligned}
\mathcal{F}^k = \Big(x \leftarrow G(\beta_i^v) \ : \ y_i^v = k, \forall i \in \{1..N_v\}, \forall v \in \{1..M\} \Big) \\
s.t. \  P(k|\mathcal{F}^k_{t-1}) \ge P(k|\mathcal{F}^k_{t})
\end{aligned}$}
\end{equation}

\noindent be all the sampled frames from the distributions with corresponding class $k$, ordered according their softmax scores. We select the top $T$ frames in $\mathcal{F}^k$ for training:

\begin{equation}
\label{eq:clTraining}
\big( \mathcal{F}_t^k \big)_{t=1}^{T} \ : \ T = h|\mathcal{F}^k|, \ h \in [0,1]
\end{equation}

\noindent With this approach, we select the frames where the classifier is most confident, which amounts to selecting the most relevant frames for each class within the plateaus. 
Note that Equation~\ref{eq:clTraining} ranks frames from all videos, and thus is independent of the number of action repetitions in one video.
While with this strategy we feed the classifier fewer noisy samples, we are still potentially missing relevant frames outside the initial plateaus. 
After training the base model, we proceed to update the sampling distributions aiming to correct misplaced plateaus so that we can feed more relevant frames. 

\begin{figure}[t!]
	\centering
	\includegraphics[width=\columnwidth]{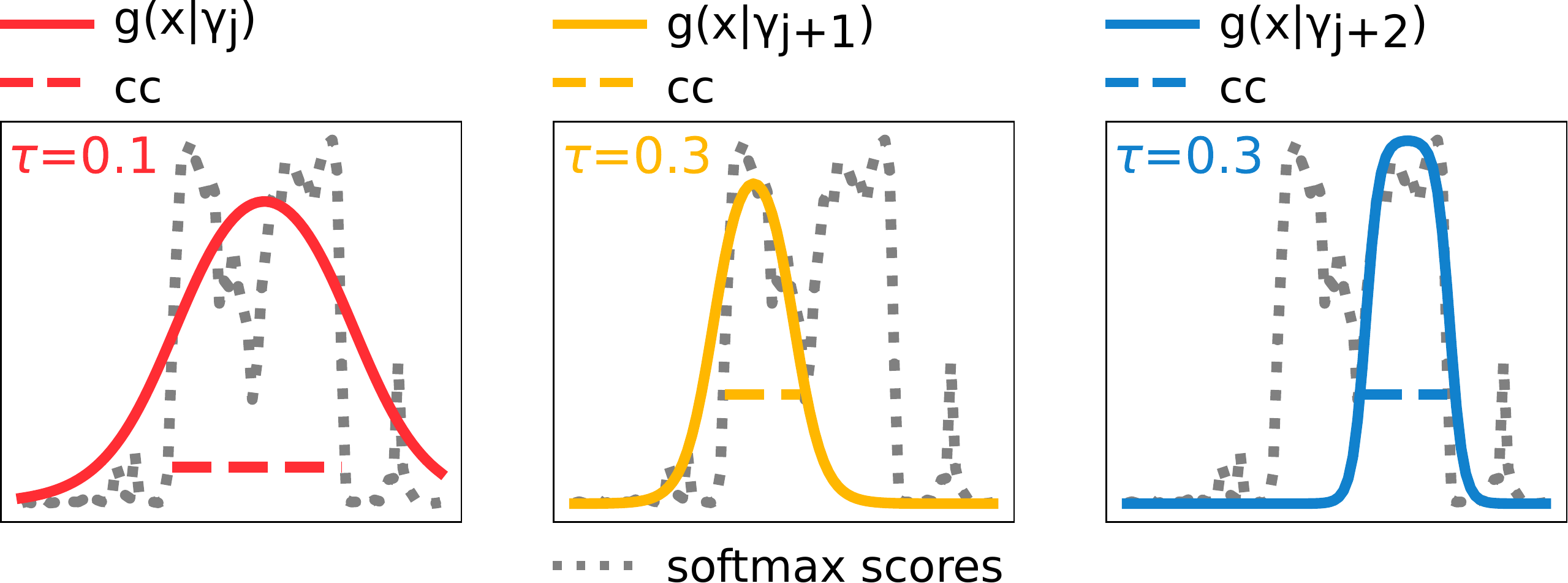}
	\caption{Finding multiple update proposals. `cc' denotes the connected components used to fit the softmax scores.}
	\label{fig:updateProposal}
\end{figure}

\subsection{Updating the Distribution Parameters}
\label{sec:update}

We assume that, overall, the initial plateaus are reasonably aligned with the actions. Under such assumptions, we iteratively update the sampling distributions parameters, reshaping and moving the initialised plateaus over more relevant frames, in order to reinforce the classifier. 
We first produce update proposals from the softmax scores, then rank the proposals to select the parameters that provide the most confident updates.

\vspace{-10pt}

\paragraph{Finding Update Proposals}

For each sampling distribution $G(\beta_i^v)$, we find update proposals given the softmax scores for the corresponding class $k = y_i^v$.
For simplicity, we describe this process for one sampling distribution and the softmax scores of its corresponding class $k$.

We fit the pdf in Equation~\ref{eq:plateauFunction} to the softmax scores at multiple positions and temporal scales.
This is done through setting a threshold $\tau \in [0,1]$ over the softmax scores, and finding all the connected components of consecutive frames with softmax scores above $\tau$. For each connected component, we fit the pdf and consider the resulting fitted parameters as one candidate for updating the sampling distribution. As $\tau$ is varied, multiple proposals at various scales can be produced.
Figure~\ref{fig:updateProposal} illustrates an example of three update proposals, where both the position and scale of the action are ambiguous, i.e. it is unclear which plateau is the best fit.

We denote each update proposal with $\gamma_j^v=(c_j^v,w_j^v,s_j^v)$. 
The set of update proposals for $\beta_i^v$ is thus: 

\begin{equation}
\mathcal{Q}_i^v = \Big\{\gamma_j^v \ :\ c_{i-1}^v < c_j^v < c_{i+1}^v\Big\}    
\end{equation}

\noindent Note that the constraint ${c_{i-1}^v < c_j^v < c_{i+1}^v}$ enforces the order of the actions in $v$ to be respected. 

\begin{figure*}[t]
	\centering
	\includegraphics[width=\textwidth]{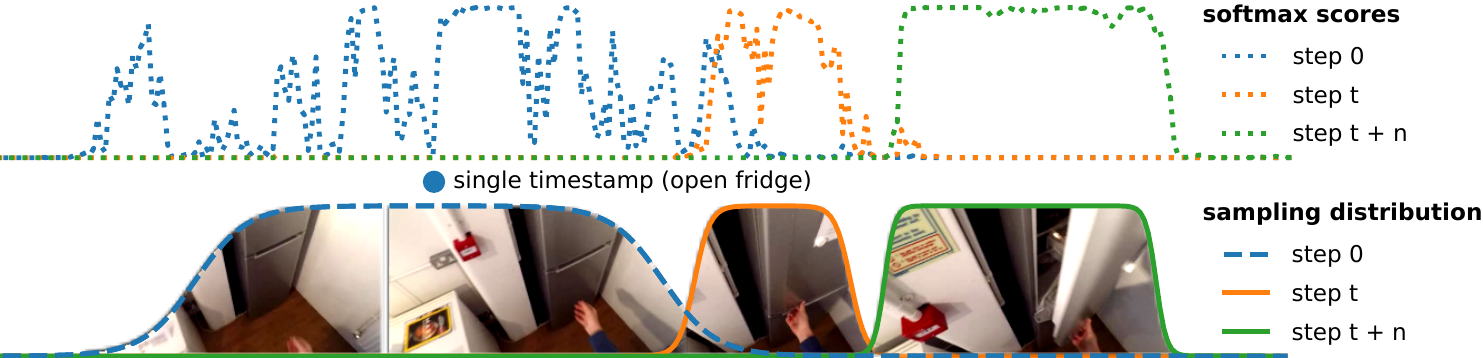}
	\caption{Updating the sampling distribution using the classifier response - example from action `open fridge' in EPIC Kitchens~\cite{Damen2018EPICKITCHENS}. Different colours indicate different training iterations.}
	\label{fig:updateExample}
\end{figure*}

\vspace{-10pt}

\paragraph{Selecting the Update Proposals}

We first define the score $\rho$ for a given plateau function $g(x|\beta_i^v)$ by averaging the softmax scores enclosed by the plateau as follows. Let $\mathcal{X}$ be the set of frames such that ${\mathcal{X} = \{x\ : \ g(x|\beta_i^v) > 0.5\}}$. The score is then defined as follows:
\begin{equation}
\rho(\beta_i^v) = \frac{1}{|\mathcal{X}|} \sum_{x \in \mathcal{X}} P(y_i^v|x)
\end{equation}

\noindent We define the confidence $\psi$ of each proposal $\gamma_j^v \in \mathcal{Q}_i^v$  as:
\begin{equation}
\psi(\gamma_j^v) = \rho(\gamma_j^v) - \rho(\beta_i^v)
\label{eq:confidence}
\end{equation}
The underlying idea is to reward proposals whose plateaus contain frames that, on average, are scoring higher than those contained within the plateau to be updated, and thus are likely to be more relevant to the action.
Accordingly, we discard update proposals with nonpositive confidence.
We select the proposal $\widehat{\gamma_i^v}$ with highest confidence for each $\beta_i^v$:

\begin{equation}
\widehat{\gamma_i^v} = \underset{\gamma_j^v}{\arg\max} \ \psi(\gamma_j^v) \ : \ \gamma_j^v \in \mathcal{Q}_i^v
\end{equation}

\paragraph{Updating Proposals}

We adopt a curriculum learning paradigm for the update as well, updating only distributions for which the selected proposals have high scores. Let:
\begin{equation}
\begin{split}
\Gamma = \Big( \widehat{\gamma_i^v} \ ,  \forall i \in \{1..N_v\}, \forall v \in \{1..M\} \Big) \\
s.t. \ \psi(\Gamma_{t-1}) \ge \psi(\Gamma_t)
\end{split}
\end{equation}
\noindent be the sequence of all selected update proposals ordered according their confidence. We pick the top $R$ proposals in $\Gamma$ to update the corresponding sampling distributions:
\begin{equation}
\label{eq:clUpdate}
\Gamma^R = \big( \Gamma_t \big)_{t=1}^{R} \ : \ R = z|\Gamma|, \ z \in [0,1]
\end{equation}
The corresponding sampling distribution parameters $\beta_i^v$ are then updated as follows:

\begin{equation}
\forall \widehat{\gamma_i^v} \in \Gamma^R \rightarrow \beta_i^v = \beta_i^v - \Lambda \Big( \beta_i^v - \widehat{\gamma_i^v} \Big)
\label{eq:update}
\end{equation}

\noindent where $\Lambda = \{\lambda_c, \lambda_w, \lambda_s\}$ denotes the set of hyperparameters controlling the velocity of the update. 
Note that we use a different update rate for the various parameters $(c, w, s)$:

\begin{equation}
c_i^v = c_i^v - \lambda_c \Big( c_i^v - \widehat{c_i^v} \Big)
\end{equation}

\noindent and similarly for $w_i^v$ and $s_i^v$. 
We update proposals until convergence. This is readily assessed by observing the average confidence of the selected proposals approaching 0.

Figure \ref{fig:updateExample} shows an example for updating one sampling distribution for class `open fridge'. The labelled timestamp and the corresponding initial sampling distribution (dotted blue and dashed blue lines) are not well aligned with the action, both positioned before the actual occurrence of the action. 
After a few iterations, the classifier is predicting the action with more confidence over frames located outside the initial plateau (dotted orange, top). The final sampling distribution (solid green, bottom) successfully aligns with the frames of the subject opening the fridge.

\begin{figure}[t!]
	\centering
	\includegraphics[width=0.9\columnwidth]{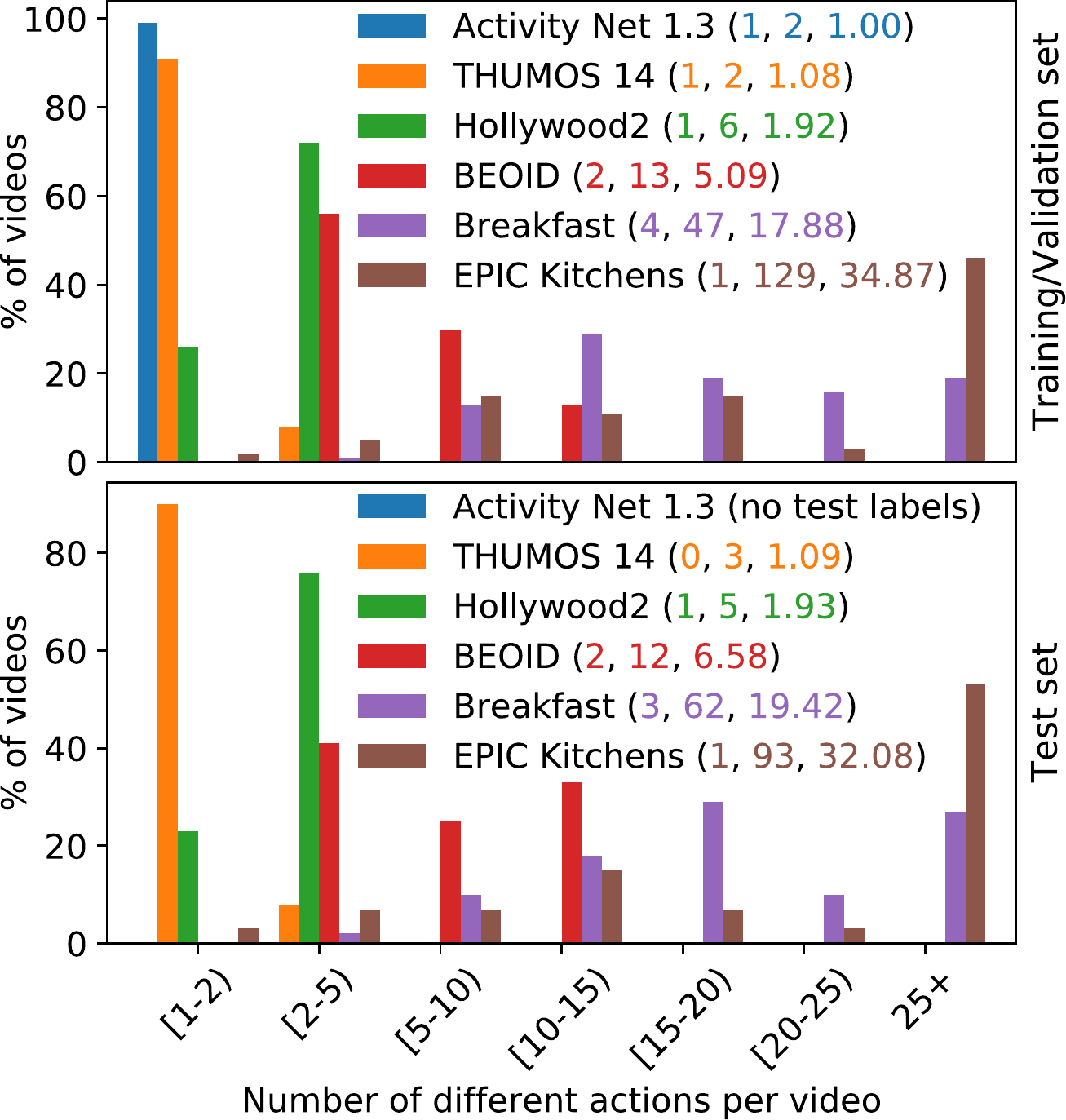}
	\caption{Different actions per video for various datasets. Numbers between parenthesis indicate (min, max, average) unique actions per video. For Activity Net~\cite{caba2015activitynet} both train and validation sets were considered, while for THUMOS 14~\cite{jiang2014thumos} we considered only the validation set. We used the `s1' split and fine segmentation labels for Breakfast~\cite{kuehne2014language}. For EPIC Kitchens~\cite{Damen2018EPICKITCHENS} we consider only the videos in the used subset.}
	\label{fig:classHistograms}
\end{figure}

\begin{table}[t]
\centering
\resizebox{\columnwidth}{!}{%
\begin{tabular}{@{}clcccccc@{}}
\toprule
\textbf{Set} & \textbf{Dataset} & \textbf{\begin{tabular}[c]{@{}c@{}}N. of\\ classes\end{tabular}} & \textbf{\begin{tabular}[c]{@{}c@{}}N. of\\ videos\end{tabular}} & \textbf{\begin{tabular}[c]{@{}c@{}}N. of\\ actions\end{tabular}} & \textbf{\begin{tabular}[c]{@{}c@{}}Avg video \\ length\end{tabular}} & \textbf{\begin{tabular}[c]{@{}c@{}}Avg classes \\ per video\end{tabular}} & \textbf{\begin{tabular}[c]{@{}c@{}}Avg actions\\ per video\end{tabular}} \\ \midrule
\multirow{3}{*}{\rotatebox[origin=c]{90}{Train}} & THUMOS 14 & 20 & 200 & 3003 & 208.90 & 1.08 & 15.01 \\
 & BEOID & 34 & 46 & 594 & 61.31 & 5.09 & 12.91 \\
 & EPIC Kitchens & 274 & 79 & 7060 & 477.37 & 34.87 & 89.36 \\ \midrule
\multirow{3}{*}{\rotatebox[origin=c]{90}{Test}} & THUMOS 14 & 20 & 210 & 3307 & 217.16 & 1.09 & 15.74 \\
 & BEOID & 34 & 12 & 148 & 57.78 & 6.58 & 12.33 \\
 & EPIC Kitchens & 274 & 26 & 1949 & 399.62 & 32.08 & 74.96 \\ \bottomrule
\end{tabular}%
}
\vspace{3pt}
\caption{Datasets information. Average video length is in seconds.}
\label{table:datasets}
\end{table}

\section{Experiments}
\label{sec:experiments}

\subsection{Datasets}
\label{sec:datasets}

Figure~\ref{fig:classHistograms} compares  various common datasets~\cite{caba2015activitynet,jiang2014thumos,marszalek09,Damen2014a,kuehne2014language,Damen2018EPICKITCHENS} for action recognition and localisation, based on the number of different actions per video in both train (top) and test (bottom) sets. The figure shows how these datasets range from an average of one action per video (Activity Net, THUMOS 14) to a maximum average of 34 actions per video (EPIC Kitchens). 
When learning from untrimmed videos with weak temporal supervision, the number of \textit{different} actions per video plays a crucial role. 
We thus evaluate our method covering this spectrum by selecting three datasets with increasing number of classes per video, namely THUMOS 14~\cite{jiang2014thumos}, BEOID~\cite{Damen2014a} and EPIC Kitchens~\cite{Damen2018EPICKITCHENS}. We show in Section~\ref{sec:results} that, as the number of different actions per video increases, video-level labels no longer provide sufficient temporal supervision, while single timestamps constitute a valid compromise between annotation effort and accuracy. 

For THUMOS 14 we use the subset of videos that were temporally labelled for 20 classes, while for BEOID we randomly split the untrimmed videos in an 80-20\% proportion for training and testing. For EPIC Kitchens we use a subset of the dataset selecting participants P03, P04, P08 and P22. With a total of 13.5 hours footage length this subset amounts to 25\% of full the dataset. 
Table~\ref{table:datasets} summarises various statistics of the chosen datasets. 
Despite considering a subset of the full dataset, EPIC Kitchens is by far the most challenging, given its very long videos containing many different actions. Additionally, EPIC-Kitchens offers novel narration annotations, as we discuss in Section~\ref{sec:singleTimestamps}.

\subsection{Implementation Details} 

We use the Inception architecture with Batch Normalisation (BN-Inception)~\cite{normalization2015accelerating} pre-trained on Kinetics~\cite{carreira2017quo}, and use TV-L1 optical flow images~\cite{zach2007duality}, with stack size 5. For training, we sample 5 stacks per action instance, and use average consensus as proposed in~\cite{wang2016temporal}. When comparing to full temporal supervision using the start/end action times, the stacks are sampled randomly within equally sized snippets, as in~\cite{wang2016temporal}. 
For faster evaluation, we uniformly sample 10 stacks from the trimmed test videos and take the centre crop using the average score for the final prediction. 
We use Adam Optimiser with batch size 256, fixed learning rate equal to $10^{-4}$, dropout equal to 0.7 and no weight decay. 

We initialise the sampling distributions with $w=45$ frames (1.5 seconds at 30 fps) and $s=0.75$ for all datasets. As we show in Section~\ref{sec:results}, our method is robust to the choice of the initial parameters. 
We train the base model for 500 epochs, to ensure a sufficient initialisation, then update the sampling distributions running the method for 500 additional epochs. The initial 500 epochs were largely sufficient for the test error to converge in all experiments before the update started. 
After training the base model with curriculum learning, we gradually increase $h$ (see Equation~\ref{eq:clTraining}) until reaching $h=1$, which corresponds to using all the sampled frames. 
We use a fixed $z=0.25$ to select the top $R$ update proposals (see Equation~\ref{eq:clUpdate}).
We vary $h$ to control noise in training frames, and keep $z$ fixed. Increasing $z$ primarily speeds the update of the distribution parameters and is akin to changing the method's learning rate.
To produce the update proposals, we use $\tau \in \{0.1, 0.2, \dots, 1\}$ and discard connected components shorter than 15 frames. 
We set update parameters $(\lambda_c, \lambda_w, \lambda_s) = (0.5, 0.25, 0.25)$ for all datasets, updating the sampling distributions every 20 epochs.
Our code uses PyTorch and is publicly available\footnote{\url{https://bitbucket.org/dmoltisanti/action_recognition_single_timestamps/}}.

\subsection{Single Timestamps}
\label{sec:singleTimestamps}
The EPIC Kitchens dataset~\cite{Damen2018EPICKITCHENS} was annotated using a two stages approach: videos were firstly narrated by the participants, through audio live narration, to produce a rough temporal location of the performed actions, from which action boundaries were then refined using crowd sourcing. We use the \textbf{narration} start timestamp as our single timestamp for training. These timestamps come from the narration audio track and exhibit a challenging offset with respect to the actions occurrences in the videos: 55.8\% of the narration timestamps are not contained in the corresponding labelled boundaries. For the timestamps outside the bounds, the maximum, average and standard deviation distance to the labelled boundaries were respectively 11.2, 1.4 and 1.6 seconds. To the best of our knowledge, this paper offers the first attempt to train for fine-grained action recognition on EPIC Kitchens using only the narration timestamps.

THUMOS 14 and BEOID do not have single timestamp annotations. We simulate \textit{rough} single timestamps from the available labels, drawing each $a_i$ from the \textit{uniform} distribution $[\sigma_i - 1sec, \epsilon_i + 1sec]$, where $\sigma_i$ and $\epsilon_i$ denote the labelled start and end times of action $i$. This approximately simulates the same live commentary annotation approach of EPIC Kitchens. We refer to this set of annotations as \textbf{TS}.

We also use another set of single timestamps for all the three datasets, where each $a_i$ is sampled using a normal distribution with mean $\frac{\sigma_i+\epsilon_i}{2}$ and standard deviation $1sec$.  
This assumes that annotators are likely to select a point close to the middle of the action when asked to provide only one timestamp.
We refer to this second set of points as \textbf{TS~in~GT}.

\begin{figure*}[t!]
	\centering
	\includegraphics[width=\textwidth]{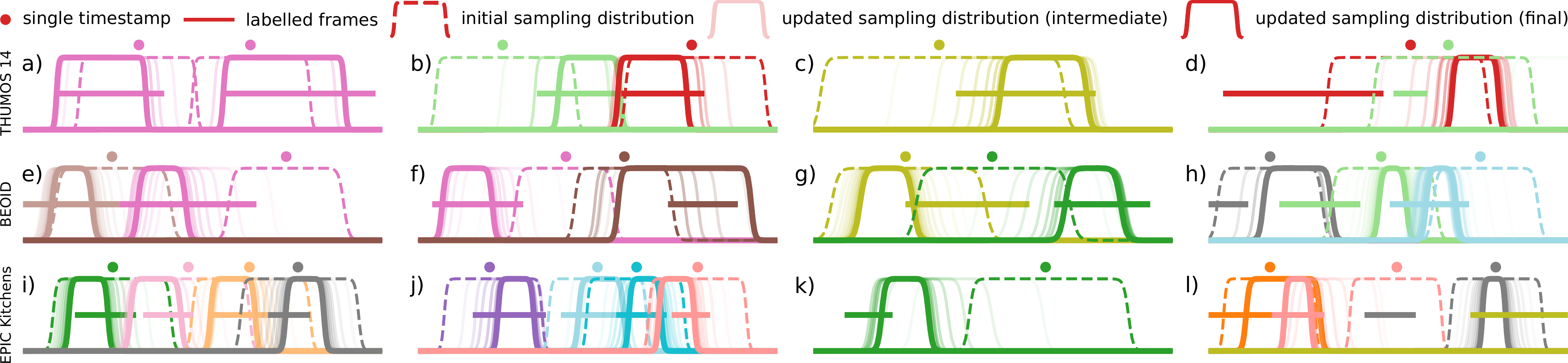}
	\caption{Qualitative results on the three datasets, plotted from results obtained with CL $h=0.50$. Different colours indicate different classes on a per-dataset basis. Labelled frames used only for plotting. Video with class labels in supplementary material.}
	\label{fig:qualitativeResults}
\end{figure*}

\subsection{Results}
\label{sec:results}

The evaluation metric used for all experiments is {top-1} accuracy. We first evaluate the \textbf{TS} timestamps with curriculum learning (CL) for training the base model running experiments with ${h \in \{0.25, 0.50, 0.75\}}$, as well as using all the sampled frames for training ($h = 1$).

\begin{table}[t]
\resizebox{\columnwidth}{!}{%
\begin{tabular}{@{}lcccc}
\toprule
\textbf{Dataset} & \textbf{CL \textit{h}} & \textbf{Before update} & \textbf{After update} & \multicolumn{1}{c}{} \\ \midrule
\multirow{4}{*}{THUMOS 14} & 0.25 & 26.10 & 28.88 \\
 & 0.50 & 32.69 & 55.15 & \multicolumn{1}{c}{} \\
 & 0.75 & 33.59 & 56.42 & \multicolumn{1}{c}{} \\
 & \small{1.00} & 63.41 & 63.53 & \multicolumn{1}{c}{} \\ \midrule
\multirow{4}{*}{BEOID} & 0.25 & 47.97 & 52.70 \\
 & 0.50  & 71.62 & 83.11 &  \\
 & 0.75  & 74.32 & 83.11 &  \\
 & \small{1.00} & 64.86 & 70.27 &  \\ \midrule
\multirow{4}{*}{EPIC Kitchens} & 0.25 & 20.47 & 22.83 \\
 & 0.50 & 21.39 & 25.35  &  \\
 & 0.75 & 20.73 & 23.86 &  \\
 & \small{1.00} & 23.55 & 24.17 & \\ \bottomrule
\end{tabular}%
}
\vspace{3pt}
\caption{Top-1 accuracy obtained with single timestamp supervision on the \textbf{TS} point set. CL \textit{h} indicates the \textit{h} parameter used for training the base model (see Equation~\ref{eq:clTraining}).}
\label{table:mainResults}
\end{table}

\begin{figure}[t!]
	\centering
	\includegraphics[width=0.9\columnwidth]{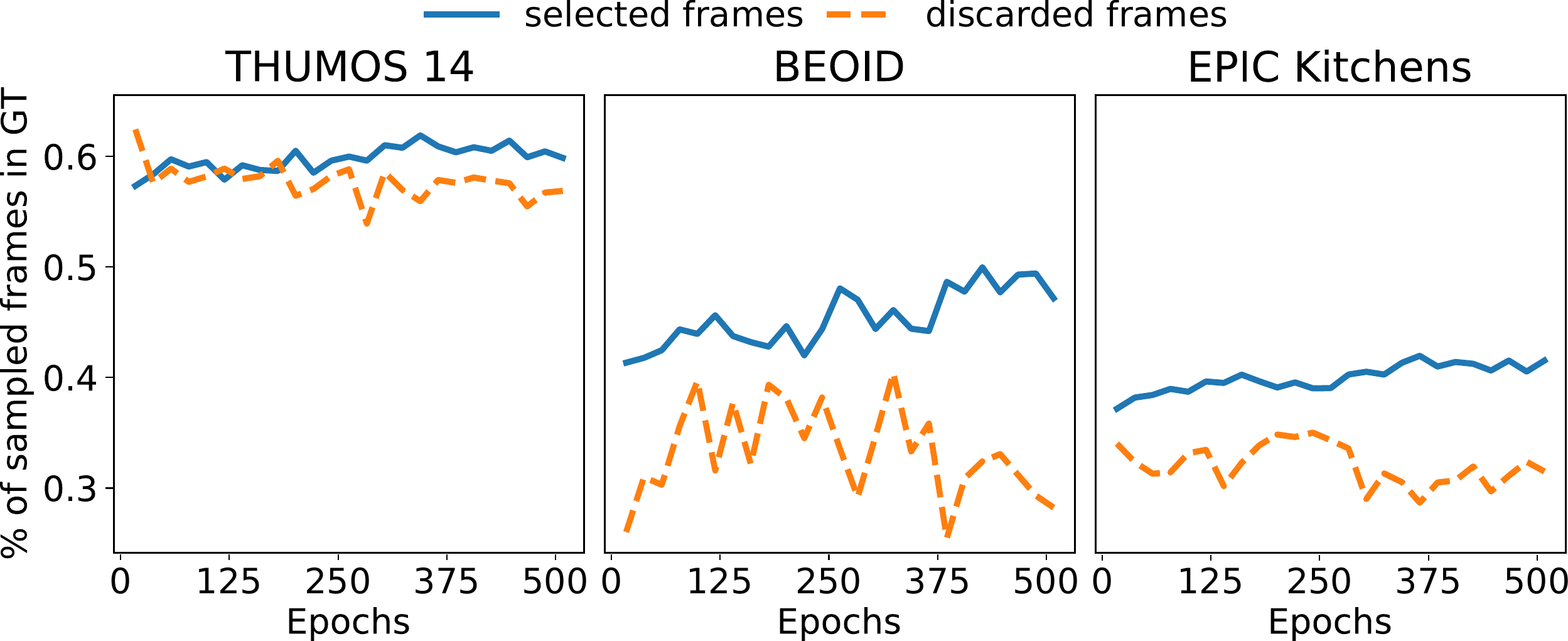}
	\caption{Percentage of sampled framed contained within labelled bounds, over training epochs (CL $h=0.50$, before update).}
	\label{fig:pointsInGt}
\end{figure}

As shown in Table~\ref{table:mainResults}, results obtained after the update consistently outperform those obtained before the update, for all datasets and for all $h$ values. 
For BEOID and EPIC, our CL strategy reduces the amount of noisy frames when training the base model, i.e. the best results are obtained with $h=0.50$. 
However, on THUMOS 14, the CL approach for the base model is less effective, with the best performance achieved when all frames are used in training.
We further analyse this in Figure~\ref{fig:pointsInGt}, which illustrates the percentage of selected and discarded frames that were enclosed by the labelled action boundaries (used only for plotting), before update. 
For BEOID and EPIC Kitchens, we notice a neat separation between the selected and discarded frames. This shows that the CL strategy was effectively picking the most relevant frames within the plateaus during training. For THUMOS 14, we do not observe the same distinct trend. A balance between the plateau width and the number of sampled frames might resolve this, but we leave this for future work. 

\begin{figure}[t!]
	\centering
	\includegraphics[width=\columnwidth]{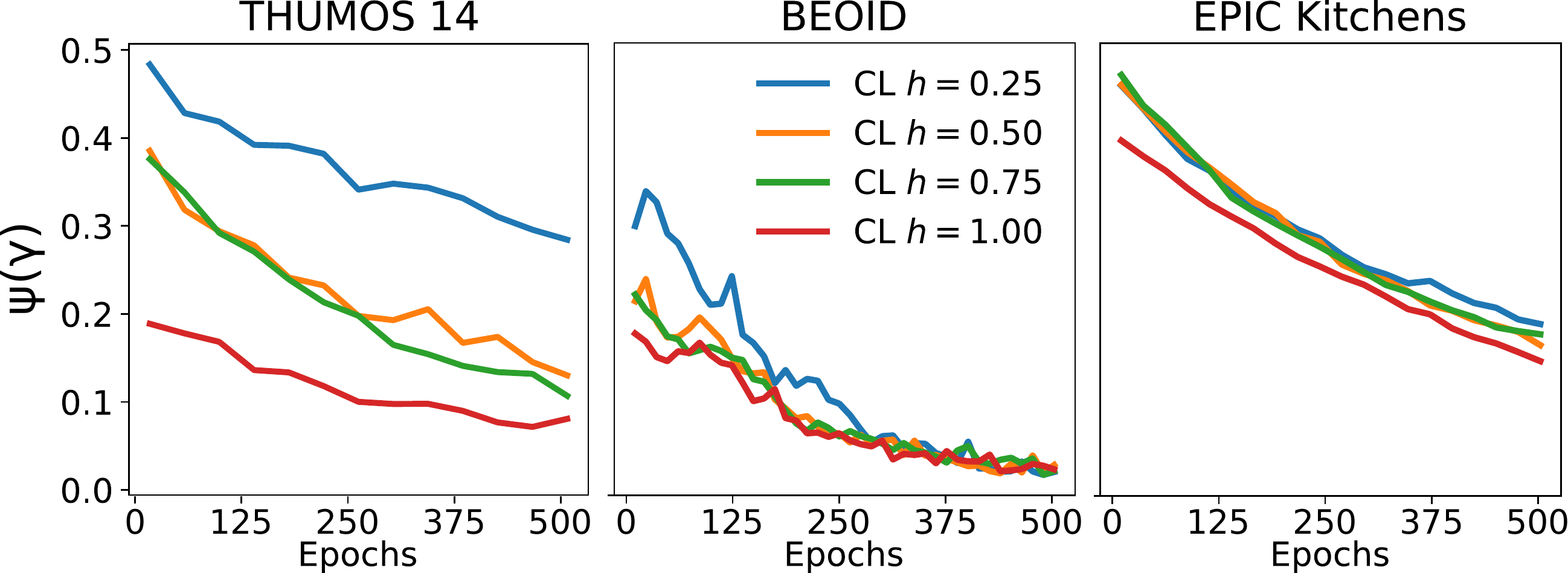}
	\caption{Average confidence of selected update proposals, as calculated in Equation~\ref{eq:confidence}, over training epochs.}
	\label{fig:convergence}
\end{figure}

In Figure~\ref{fig:convergence}, we assess the update convergence by plotting the average confidence of the selected update proposals over training epochs. For all cases, the average confidence decreases steadily, indicating the classifier's convergence.

We illustrate a few examples from each dataset in Figure~\ref{fig:qualitativeResults}, showing the iterative update of the sampling distributions. The examples are plotted from results obtained with CL $h=0.50$ on the \textbf{TS} point set. Our update method is able to successfully refine the sampling distributions even when the initial plateaus are considerably overlapping with other unrelated actions (subplots \textit{e, g, i, j}) or when the initial plateaus contain much background (subplots \textit{b, c, e, f, k}). We highlight a few failure cases as well. In subplots \textit{g} (light green plateau) and \textit{h} (grey plateau), the initial plateaus are pushed outside the relevant frames. In both cases, the number of training examples was small (8 and 5 instances), with the single timestamps located almost always outside the action. 
In subplot \textit{l}, the pink and grey initial plateaus were shifted with respect to the corresponding actions, reflecting the challenge EPIC Kitchens poses when using narration timestamps. While the update method managed to recover the correct location for the pink plateau, the grey plateau did not converge to the relevant frames.

\vspace{-10pt}

\begin{figure}[t!]
	\centering
	\includegraphics[width=\columnwidth]{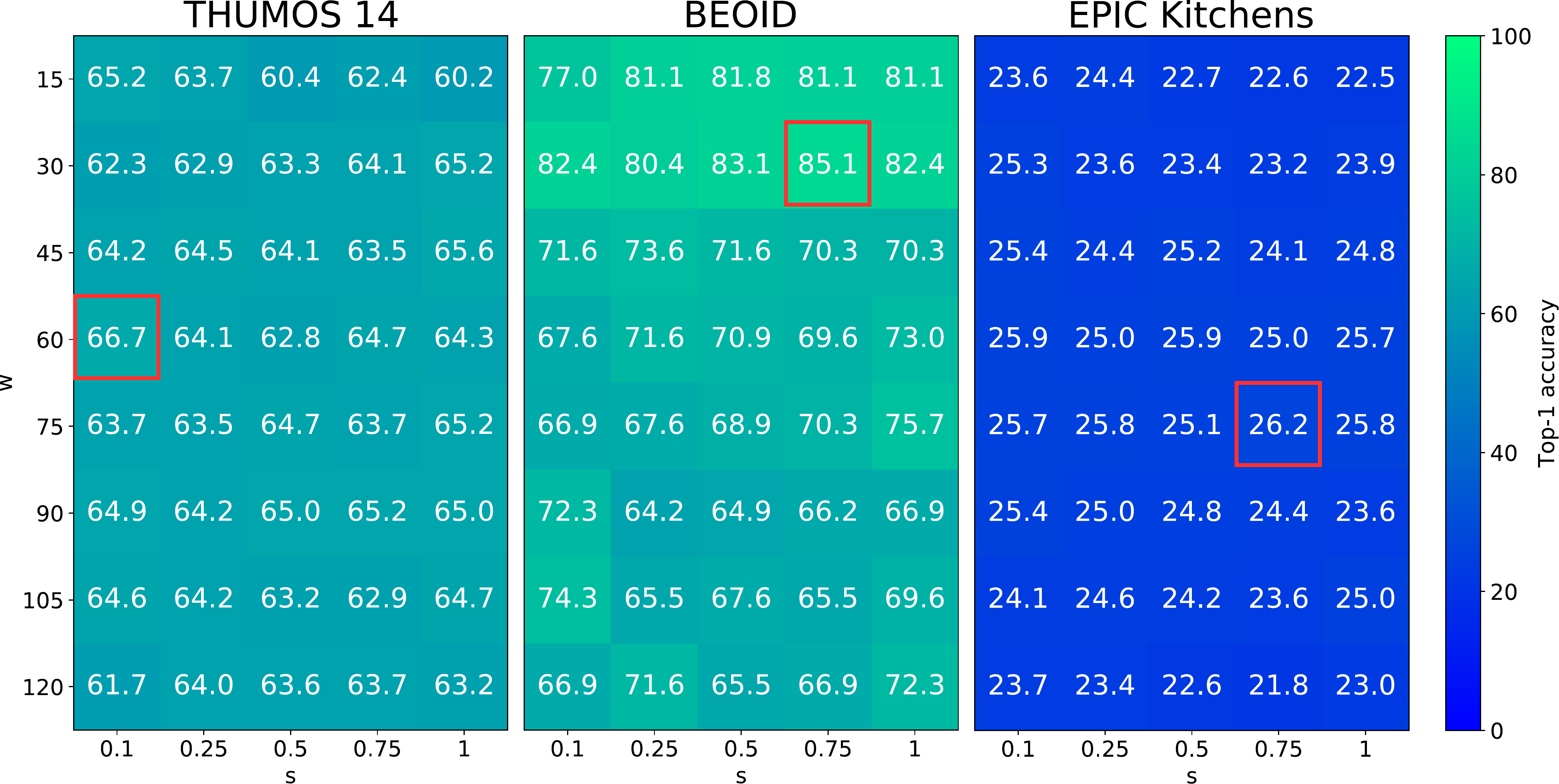}
	\caption{Top-1 accuracy obtained after update with different initial $w$ and $s$, with CL $h=1$. Red boxes highlight best results.}
	\vspace{-5pt}
	\label{fig:gridSearch}
\end{figure}

\paragraph{Parameters initialisation} We assess the impact of the initial parameters $w$ and $s$ for the sampling distributions through a grid search. Figure~\ref{fig:gridSearch} compares top-1 accuracy obtained after update with different $(w,s)$ combinations, using CL $h = 1.00$.
We observe that for the two large datasets (THUMOS 14 and EPIC Kitchens), our method is robust to the initialisation of both $w$ and $s$, i.e.\ similar performance is obtained for all parameters combinations. Decreased robustness for BEOID is potentially due to the small size of the dataset.

We note that the best results obtained via the grid search (highlighted with red boxes in the Figure) are slightly superior to those previously reported  in Table~\ref{table:mainResults}. This is because when the plateaus are optimally initialised, we are less likely to sample noisy frames when training. 

\subsection{Comparing Levels of Supervision}

We now compare different levels of temporal supervision, namely the weakest video-level labels, single timestamps (both \textbf{TS} and \textbf{TS in GT} point sets) and full temporal boundaries. 
Particularly, we show that video-level supervision, while being the least expensive to gather, cannot provide a sufficient supervision when dealing with videos containing multiple different actions. 

We choose Untrimmed Net~\cite{wang2017untrimmednets} amongst the aforementioned works~\cite{wang2017untrimmednets,nguyen2017weakly,shou2018autoloc,paul2018w}, which is used to extract features in~\cite{paul2018w} and is the backbone model of~\cite{shou2018autoloc}, due to the availability of published code. 
We train Untrimmed Net using uniform sampling and hard selection module, using the same BN-Inception architecture and Kinetics pre-trained weights used for our baselines. For Untrimmed Net we report results obtained on RGB images as these performed better than flow images in all our experiments.

Table~\ref{table:untrimmedComparison} compares the results obtained with the three temporal supervisions. Single timestamps results are reported after update, with CL $h = 1.00$.   
When only one class of action is contained in the videos, as in THUMOS 14, Untrimmed Net notably achieves virtually the same results as the fully supervised baseline. However, as the average number of different actions per video increases, it becomes increasingly harder for video-level supervision to achieve sufficient accuracy. 
In~\cite{wang2017untrimmednets} when a video contains action instances from multiple classes, the label vector is $L^1$-normalised so that all the present classes contribute equally to the cross-entropy loss. As a consequence, without any temporal labels, it is very hard to train the model when a large number of classes are present in a video. 

Results obtained with single timestamps remain comparable to full supervision for all datasets, though requiring significantly less labelling effort\footnote{For completion, accuracy before update for TS was 64.74, 73.65 and 25.19 for THUMOS 14, BEOID and EPIC Kitchens. For TS in GT, accuracy before update was 64.74, 85.81 and 31.66.}. For THUMOS~14 and BEOID, we observe little difference between the point sets \textbf{TS} and \textbf{TS in GT}. 
For EPIC Kitchens, which has the largest number of distinct classes per video, we notice a larger gap in performance with respect to the fully supervised baseline. However, when drawing the initial timestamps from the labelled bounds (\textbf{TS in GT}), we achieve higher accuracy. 
From these results we conclude that single timestamps supervision constitutes a good compromise between accuracy and annotation effort. 

\begin{table}[t]
\resizebox{\columnwidth}{!}{%
\begin{tabular}{@{}lccccc@{}}
\toprule
\textbf{Baseline} & & U. Net\cite{wang2017untrimmednets} & \multicolumn{3}{c}{Ours} \\ \midrule
\textbf{Supervision} & \textbf{\textcolor[rgb]{0.2,0.4,0.8}{APV}} & Video-level & TS & TS in GT & Full \\ \midrule
THUMOS 14 & \textcolor[rgb]{0.2,0.4,0.8}{1.08} & 64.92 & 66.68 & 64.53 & 67.10 \\ 
BEOID & \textcolor[rgb]{0.2,0.4,0.8}{5.09} & 28.37 & 85.14 & 88.51 & 87.83 \\
EPIC Kitchens & \textcolor[rgb]{0.2,0.4,0.8}{34.87} & 2.20 & 26.22 & 32.53 & 35.97 \\ \bottomrule
\end{tabular}
}
\vspace{3pt}
\caption{Comparison between different levels of temporal supervision. \textcolor[rgb]{0.2,0.4,0.8}{APV} indicates the average number of unique actions per training video. TS results refer to the accuracy obtained with the best initialisation (see Figure~\ref{fig:gridSearch}). Timestamp results are reported after update, with $h = 1.00$.}
\label{table:untrimmedComparison}
\end{table}

\begin{table}[t]
\resizebox{\columnwidth}{!}{%
\begin{tabular}{@{}lccccc@{}}
\toprule
\textbf{Baseline} & \textbf{mAP@0.1} & \textbf{mAP@0.2} & \textbf{mAP@0.3} & \textbf{mAP@0.4} & \textbf{mAP@0.5} \\ \midrule
Ours (Full) & 26.7 & 22.5 & 18.5 & 14.3 & 11.1 \\
Ours (TS) & 24.3 & 19.9 & 15.9 & 12.5 & 9.0 \\ \midrule
U. Net~\cite{wang2017untrimmednets} & 44.4 & 37.7 & 28.2 & 21.1 & 13.7 \\ \bottomrule
\end{tabular}
}
\vspace{3pt}
\caption{Localisation results on THUMOS 14 at different IoUs.}
\vspace{-10pt}
\label{table:localisation}
\end{table}

\vspace{-5pt}
\subsection{Future Direction: Localisation with TS} 
\vspace{-5pt}

In this work we focus on single timestamp supervision for action classification. Using solely frame-level classification scores to localise the extent of actions would be sub-optimal. We show this in Table~\ref{table:localisation}, which presents mean average precision (mAP) on THUMOS 14 obtained with our baselines, compared to~\cite{wang2017untrimmednets}. 
We follow the localisation pipeline of~\cite{wang2017untrimmednets}, fusing RGB and flow scores obtained with full and single timestamp (TS) supervision. While TS performs comparably to full supervision, even full supervision is inferior to~\cite{wang2017untrimmednets}, which is optimised for localisation. Our approach could be extended to localisation by supervising a temporal model (e.g. RNN) from plateau functions to learn the temporal boundaries. We leave this for future work.

\vspace{-5pt}
\section{Conclusions}
\vspace{-5pt}

In this work we investigate using single timestamp supervision for training multi-class action recognition from untrimmed videos. We propose a method that initialises and iteratively updates sampling distributions to select relevant training frames, using the classifier's response. 
We test our approach on three datasets, with increasing number of unique action classes in training videos.
We show that, compared to video-level supervision, our method is able to converge to the locations and extents of action instances, using only single timestamp supervision.
Results also demonstrate that, despite using a much less burdensome annotation effort, we are able to achieve comparable results to those obtained with full, expensive, temporal supervision. 
Extending these annotations to other tasks such as localisation is left for future work.
Future directions also include updating the sampling distribution parameters in an end-to-end differentiable manner.

\vspace{-10pt}

\paragraph{Acknowledgement} Research supported by EPSRC LOCATE (EP/N033779/1) and EPSRC DTP. We use publicly available datasets, and publish our code.

{\small
\bibliographystyle{ieee_fullname}
\bibliography{references}
}

\end{document}